%% file: main.tex
\documentclass[runningheads]{llncs}
\usepackage[T1]{fontenc}
\usepackage{graphicx}
\usepackage{amsmath}        %
\usepackage{amssymb}
\usepackage{xcolor}         %
\usepackage{booktabs}       %
\usepackage{array}          %
\usepackage{makecell}       %
\usepackage{pifont}        %
\usepackage{url}           %
\usepackage{hyperref}
\usepackage[capitalize]{cleveref}
\usepackage{enumitem}
\usepackage{multirow}
\usepackage{tabularx} %
\newcolumntype{C}{>{\centering\arraybackslash}X}

\input{macros.tex}

\graphicspath{{figures}}

\begin{document}
\title{Room Envelopes: A Synthetic Dataset for\\Indoor Layout Reconstruction from Images}
\titlerunning{Room Envelopes}
\author{
  Sam Bahrami\and
  Dylan Campbell
}

\authorrunning{S. Bahrami and D. Campbell}

\institute{
  The Australian National University, Canberra, Australia
  \email{\{sam.bahrami,dylan.campbell\}@anu.edu.au}
}
\maketitle              %
\begin{abstract}

Modern scene reconstruction methods are able to accurately recover 3D surfaces that are visible in one or more images.
However, this leads to incomplete reconstructions, missing all occluded surfaces.
While much progress has been made on reconstructing entire objects given partial observations using generative models, the structural elements of a scene, like the walls, floors and ceilings, have received less attention.
We argue that these scene elements should be relatively easy to predict, since they are typically planar, repetitive and simple, and so less costly approaches may be suitable.
In this work, we present a synthetic dataset---Room Envelopes\footnote{Project website: \url{https://sambahrami.github.io/room_envelopes}}---that facilitates progress on this task by providing a set of RGB images and two associated pointmaps for each image: one capturing the visible surface and one capturing the first surface once fittings and fixtures are removed, that is, the structural layout.
As we show, this enables direct supervision for feed-forward monocular geometry estimators that predict both the first visible surface and the first layout surface.
This confers an understanding of the scene's extent, as well as the shape and location of its objects.


\end{abstract}
\begin{figure*}[!t]
    \centering
    \begin{tabular}{c}
    \includegraphics[width=0.7\textwidth, trim=150pt 150pt 150pt 75pt]{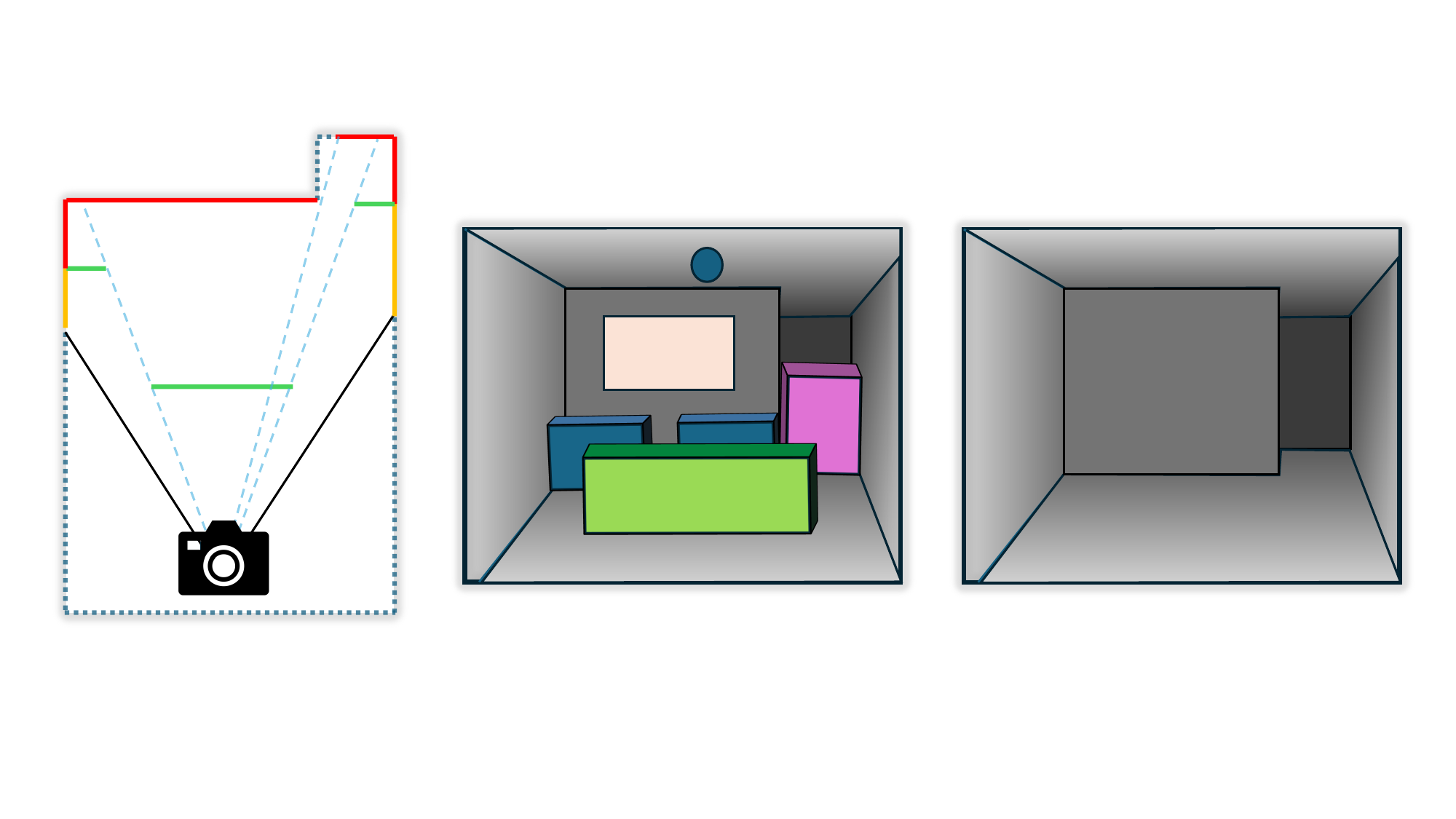} \\
    \\
    \begin{tabular}{ccc}
    (a) Floor Plan View & (b) Visible Surface & (c)  Layout Surface \\
    \\
    \includegraphics[width=0.3\textwidth]{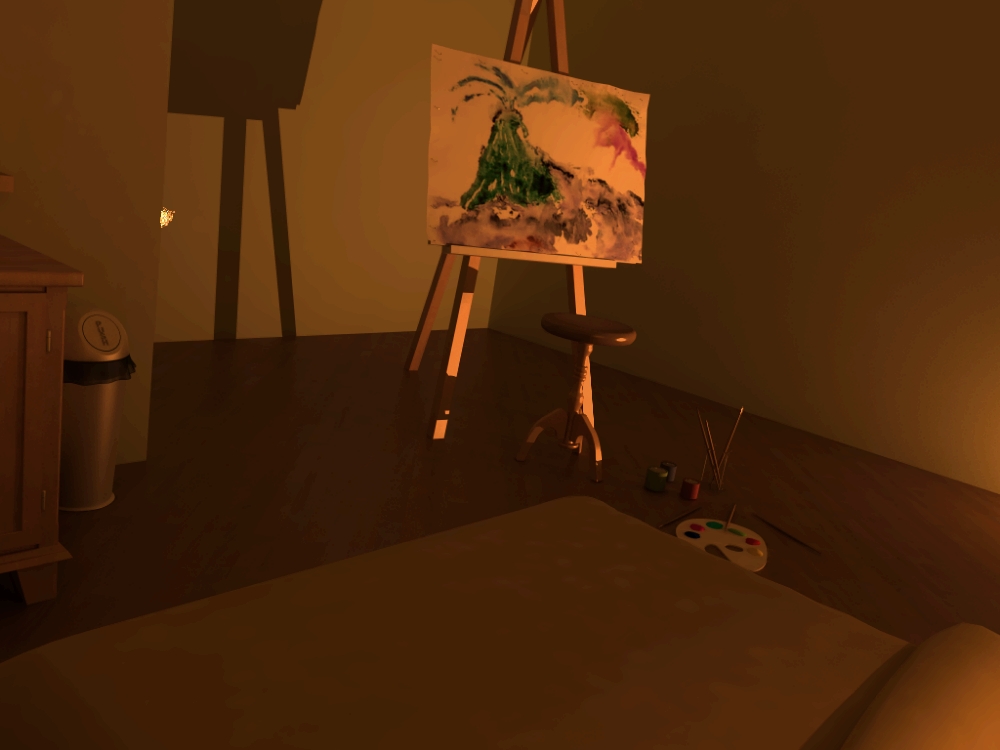} & 
    \includegraphics[width=0.3\textwidth]{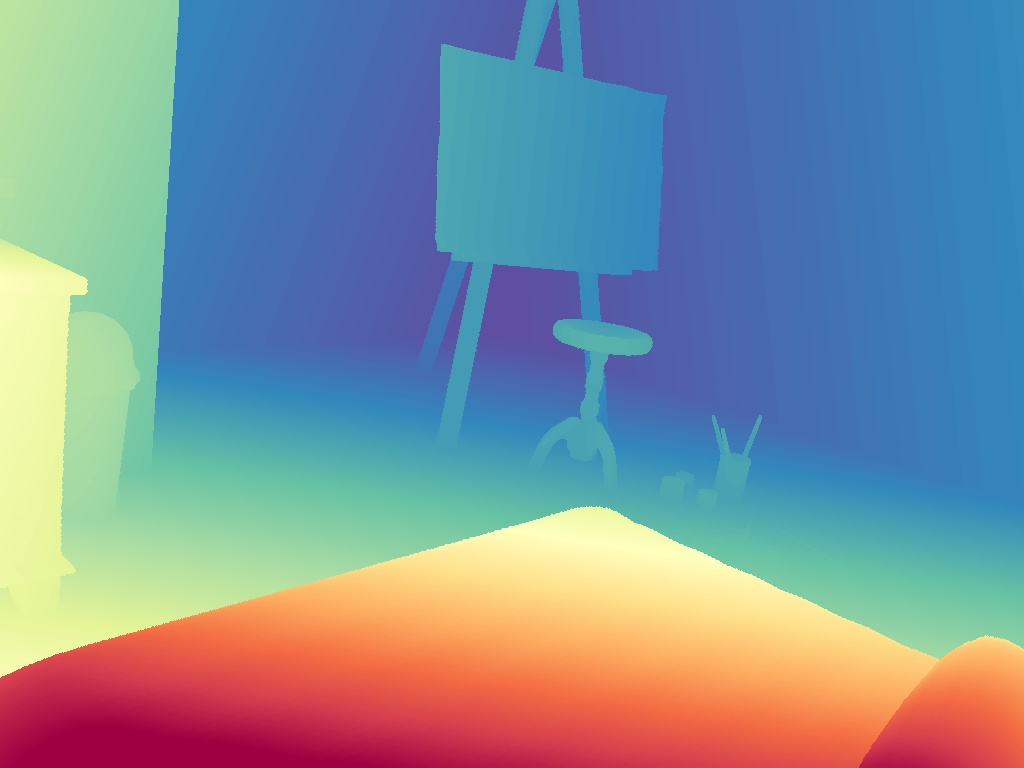} & 
    \includegraphics[width=0.3\textwidth]{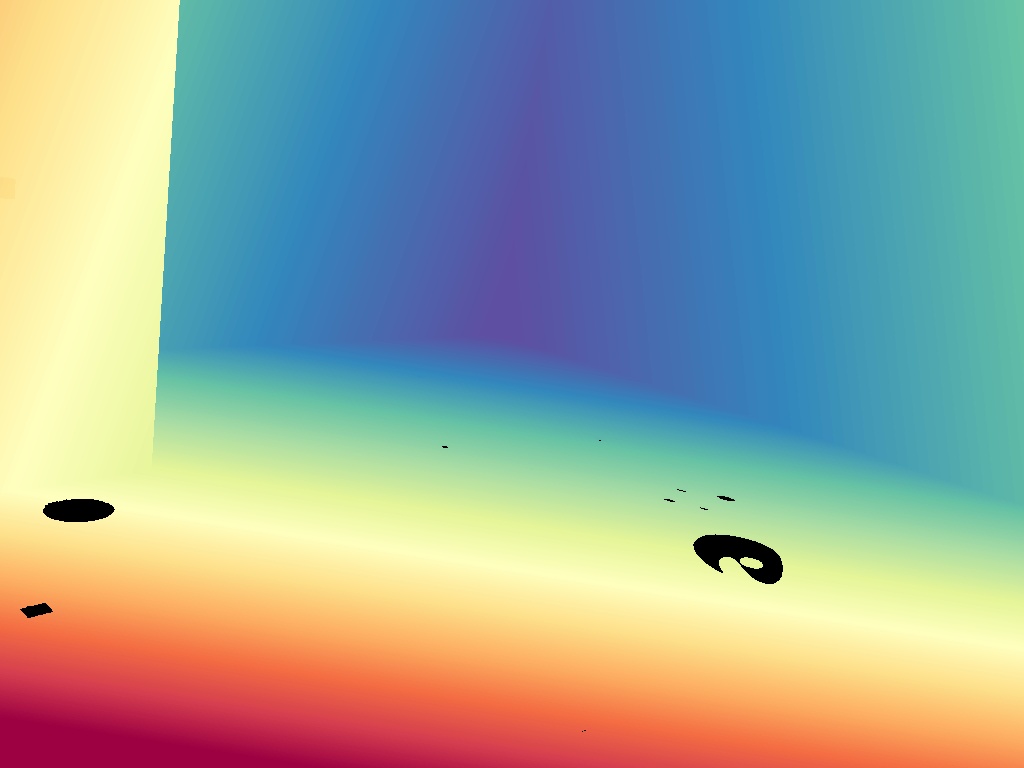} \\
    (d) Original RGB Image & (e) Visible Surface Depth & (f) Layout Surface Depth\\
    \end{tabular}
    \end{tabular}
    \vspace{-4pt}
    \caption{Room Envelopes dataset overview. Our synthetic dataset provides dual pointmap representations for indoor scene reconstruction. (a)~Overhead floor plan showing the view from a camera capturing layout areas in red, the first visible surface in green, and parts shared in both in orange. (b)~The visible surface capturing all directly visible geometry including furniture and objects.
    (c)~The layout surface showing structural elements (walls, floors, ceilings, windows, doors) as they would appear without occlusion. This dual representation enables direct supervision for layout reconstruction in occluded regions.
    (d--f)~Example data from our dataset.
    (d)~The original RGB image from Hypersim.
    (e)~The visible surface depth capturing all visible surfaces including furniture and objects
    (f)~The first layout surface depth showing only structural elements (walls, floors, ceiling, windows, doors).
    }
    \label{fig:splash}
\end{figure*}
\section{Introduction}

Indoor scene reconstruction remains a challenging problem, particularly when attempting to understand the complete spatial structure of indoor environments from limited visual information.
While recent advances in monocular depth estimation and 3D reconstruction have shown promising results, existing approaches often struggle with occluded regions and fail to provide comprehensive geometric understanding of indoor scenes outside of the first visible surface.

Current models used for indoor scene reconstruction typically use depth images \cite{wang_moge_2024,leroy_mast3r_2024} or layered depth representations \cite{li2025lari} for training, but these formats have inherent limitations.
Depth maps only capture the first visible surface, losing information about the underlying room structure that might be occluded.
Layered depth approaches attempt to address occlusion by representing multiple depth layers per pixel, but surfaces that are spatially continuous in 3D space often become fragmented across different depth layers, creating artificial discontinuities and requiring methods that perform layer assignment as well as geometry estimation.
For a given viewpoint, we refer to the union of the first visible surface and the first structural layout surface as the \textit{envelope} of the view, as illustrated in \cref{fig:splash}.
The structural layout comprises structural elements like walls, floors, ceilings, windows and doors.
This is essentially the room with fittings (removable items like furniture, appliances, curtains, carpets) and fixtures (attached items like sinks, baths, lights) removed, which largely coincides with the concept of layout in existing layout estimation works \cite{jiang2020peekaboo,mathew_layout_2020,meng20253dlayout}.
Understanding this structural layout is useful for many applications including robotic navigation, augmented reality, and architectural analysis.
However, in typical indoor scenes, this layout is often partially occluded by furniture, making it difficult for models to learn robust representations of the underlying room geometry.

Feed-forward 3D reconstruction models face particular challenges at occlusion edges, where they tend to predict averaged depth values rather than making decisive geometric predictions \cite{szymanowicz_flash3d_2024,wang_moge_2024,leroy_mast3r_2024}.
This occurs because these models optimise a reconstruction loss by predicting the statistical mean in ambiguous regions, resulting in blurry or noisy outputs at object boundaries.
Room layouts however are predominantly continuous and planar, making them well-suited for feed-forward prediction.
By focusing on layout estimation, models can leverage the geometric regularity of architectural structures to produce sharp, accurate reconstructions even in partially occluded regions.

To address these limitations, we introduce Room Envelopes, a synthetic dataset that explicitly provides two complementary pointmaps for each RGB image:
(1) one representing the first visible surface, capturing all directly visible surfaces including fittings and fixtures, and
(2) one representing the first layout surface, capturing the scene's structural elements as they would appear from the same viewpoint if the fittings and fixtures were removed.
This dual representation provides several advantages over existing dataset formats.
First, the first layout surface provides structural clarity by offering unoccluded views of room boundaries, enabling models to learn robust representations of indoor spatial structure.
Second, it facilitates the development of scene reconstructors that have less ambiguity compared to layered depth approaches, eliminating uncertainty about layer assignment.
Finally, it allows models to be trained with direct supervision on room layout estimation, rather than relying on indirect multi-view photometric losses.

Our dataset is constructed by combining multiple viewpoints from high-quality synthetic indoor scenes, similar to how real-world datasets are collected through multi-view reconstruction approaches.
This multi-view aggregation allows us to build a complete point cloud representation of indoor environments, from which we isolate structural elements and re-render layout surfaces unoccluded.
We process these pointclouds into pointmap representations which maintain spatial correspondence with the original RGB images, enable pixel-aligned supervision, and preserve the natural image structure that existing vision neural network architectures are optimised for.
This approach produces a dataset that is fundamentally equivalent to what could be achieved with the best possible real-world data collection tools for this task, with the advantage of perfect semantic labelling, perfect camera information, and controlled lighting conditions.

We demonstrate the utility of this dataset by training a model for room layout estimation that can predict the surface envelope of an indoor scene from a single RGB image.
The main contributions of this work are
\begin{enumerate}[nosep]
    \item Room Envelopes, an indoor synthetic dataset providing an image, a visible surface pointmap, and a layout surface pointmap for each camera pose; and
    \item a feed-forward scene reconstruction model demonstrating effective room layout estimation using this dataset.
\end{enumerate}

\section{Related Work}

\subsection{Indoor Scene Datasets}

Indoor scene understanding relies on both real-world and synthetic datasets, each with distinct advantages and limitations. We provide a detailed comparison of Room Envelopes with existing indoor datasets in \cref{tab:tab_indoor_datasets}.
Real-world datasets \cite{zhou2018stereo,dai2017scannet,yeshwanth2023scannet++,chang2017matterport3d} face significant collection challenges that limit their scale and quality. 
Ground truth acquisition is difficult due to depth sensor limitations, including restricted resolution, errors with reflective surfaces, artifacts at object boundaries, and these datasets require extensive manual effort for data capture and post-processing.

Synthetic datasets provide perfect ground truth at scale with reduced manual effort. 
Recent procedural generation advances enable automatic scene layout generation \cite{infinigen_indoors}, but often lack natural spatial arrangements that characterise human-designed spaces. 
Artist-designed synthetic environments offer a compelling middle ground \cite{Structured3D,jiang2025megasynth,roberts2021hypersim,3d_front}, however visual realism remains a key consideration. 
HyperSim \cite{roberts2021hypersim} uses V-Ray rendering to achieve photorealistic imagery per view. 
Room Envelopes builds on this foundation while introducing layout surface maps for each view.
\begin{table*}[t]
    \caption
    {
    Comparison of dataset features across indoor scene datasets. We show which datasets provide camera poses, point clouds, depth maps (first depth and layout depth), mesh data, semantic labels, and surface normals. Room Envelope is the only dataset that provides both first layer and layout depth representations for comprehensive 3D reconstruction training.
    }
    \centering
    \scriptsize
    \begin{tabularx}{\textwidth}{lcccccccccc}
    \toprule
    
    Datasets & \#Scenes & Type & Camera & Point & First & Layout & Mesh & Semantic & Normals\\
     & & & Poses & Cloud & Depth & Depth & & Labels & \\
    
    \midrule
    Matterport3D~\cite{chang2017matterport3d} & 90 & Real & \textcolor{green}{\checkmark} & \textcolor{black}{\xmark} & \textcolor{green}{\checkmark} & \textcolor{black}{\xmark} & \textcolor{green}{\checkmark} & \textcolor{green}{\checkmark} & \textcolor{black}{\xmark}\\
    
    ScanNet++~\cite{yeshwanth2023scannet++} & 1,006 & Real & \textcolor{green}{\checkmark} & \textcolor{green}{\checkmark} & \textcolor{green}{\checkmark} & \textcolor{black}{\xmark} & \textcolor{green}{\checkmark} & \textcolor{green}{\checkmark} & \textcolor{black}{\xmark}\\
    
    ScanNet~\cite{dai2017scannet} & 1,513 & Real & \textcolor{green}{\checkmark} & \textcolor{black}{\xmark} & \textcolor{green}{\checkmark} & \textcolor{black}{\xmark} & \textcolor{green}{\checkmark} & \textcolor{green}{\checkmark} & \textcolor{black}{\xmark}\\
    
    ARKitScenes~\cite{baruch2021arkitscenes} & 1,661 & Real & \textcolor{green}{\checkmark} & \textcolor{green}{\checkmark} & \textcolor{green}{\checkmark} & \textcolor{black}{\xmark} & \textcolor{green}{\checkmark} & \textcolor{black}{\xmark} & \textcolor{black}{\xmark}\\
    
    RealEstate10K~\cite{zhou2018stereo} & 74K & Real & \textcolor{green}{\checkmark} & \textcolor{black}{\xmark} & \textcolor{black}{\xmark} & \textcolor{black}{\xmark} & \textcolor{black}{\xmark} & \textcolor{black}{\xmark} & \textcolor{black}{\xmark}\\

    MegaSynth~\cite{jiang2025megasynth} & 700K & Synthetic & \textcolor{green}{\checkmark} & \textcolor{black}{\xmark} & \textcolor{green}{\checkmark} & \textcolor{black}{\xmark} & \textcolor{black}{\xmark} & \textcolor{black}{\xmark} & \textcolor{black}{\xmark}\\

    Structured3D~\cite{Structured3D} & 22K & Synthetic & \textcolor{green}{\checkmark} & \textcolor{black}{\xmark} & \textcolor{green}{\checkmark} & \textcolor{black}{\xmark} & \textcolor{black}{\xmark} & \textcolor{green}{\checkmark} & \textcolor{green}{\checkmark}\\
    
    3D-Front~\cite{3d_front} & 19K & Synthetic & \textcolor{black}{\xmark} & \textcolor{black}{\xmark} & \textcolor{black}{\xmark} & \textcolor{black}{\xmark} & \textcolor{green}{\checkmark} & \textcolor{black}{\xmark} & \textcolor{black}{\xmark}\\

    HyperSim~\cite{roberts2021hypersim} & 461 & Synthetic & \textcolor{green}{\checkmark} & \textcolor{black}{\xmark} & \textcolor{green}{\checkmark} & \textcolor{black}{\xmark} & \textcolor{green}{\checkmark} & \textcolor{green}{\checkmark} & \textcolor{green}{\checkmark}\\
    
    Room Envelopes (Ours) & 461 & Synthetic & \textcolor{green}{\checkmark} & \textcolor{green}{\checkmark} & \textcolor{green}{\checkmark} & \textcolor{green}{\checkmark} & \textcolor{black}{\xmark} & \textcolor{green}{\checkmark} & \textcolor{green}{\checkmark}\\
    
    \bottomrule
    
    \end{tabularx}
    \label{tab:tab_indoor_datasets}
\end{table*}

\subsection{Scene and Layout Reconstruction}

A modern approach to scene reconstruction estimates pixel-aligned pointmaps from images, where each pixel in the output pointmap represents a 3D point in the scene \cite{zhang2025advancesfeedforward}.
The pioneering feed-forward pointmap reconstruction method DUSt3R \cite{dust3r_cvpr24} takes pairs of unposed views and estimates pointmaps in the same coordinate system, encoding their geometry and pose.
This pairwise representation can be extended to estimate geometry for large scenes with many views, including as part of a SLAM system \cite{murai2025mast3r}.
The follow-up work MASt3R \cite{leroy_mast3r_2024} improved upon this approach by introducing local feature matching and modifying the training methodology to output metric-scale representations.
MoGe \cite{wang_moge_2024} extended this paradigm to single-view pointmap estimation, with MoGe v2 \cite{wang_moge2_2025} providing metric-scale output.
LaRI \cite{li2025lari} adapted the training concept from MoGe to estimate layered pointmap representations, attempting to predict the depth of occluded surfaces with a feed-forward model.
VGGT \cite{wang2025vggt} predicts comprehensive 3D scene attributes (such as pointmaps, camera poses, and depth maps) from single or multiple images of a scene without post-processing, achieving state-of-the-art performance in feed-forward 3D reconstruction.

The goal of complete 3D scene reconstruction is to generate plausible representations of entire environments, including both visible and occluded regions, from limited input views.
Regression-based methods include 3D Gaussian Splatting \cite{kerbl20233d} feed-forward approaches \cite{szymanowicz_flash3d_2024,smart2024splatt3r,zhang2024gs}, as well as encoder--decoder approaches that create intermediate scene representations for novel view synthesis or complete geometric reconstruction \cite{jin_lvsm_2025,jiang2025rayzer,wang_cut3r_2025}. 
These methods learn by comparing the rendered predicted representation with other views of the same scene in their ground truth dataset, indirectly supervising the occluded parts of the scene. 
These methods often produce impressive visual results near the input views, however, they struggle with view extrapolation, leading to poor predictions for areas that were occluded in the source views.
In contrast, generative models are able to produce sharper and often more accurate representations of the unseen areas, but compromise speed and require significant training resources \cite{xiang_trellis_2024,szymanowicz_bolt3d_2025,liu2023zero,shi2023mvdream}. 
We suspect that generative approaches are less necessary for layout estimation than object reconstruction, since the layout is typically simple, repetitive and often planar.
Both approaches, however, require data with layout supervision, which is addressed by the Room Envelopes dataset. 

Layout estimation is the task of reconstructing the structure of an indoor scene from an RGB image \cite{mathew_layout_2020,meng20253dlayout}.
Traditional approaches often rely on Manhattan world assumptions, where every wall is at right angles to every other wall, modeling scenes as the best fit 3D box defined by simple 3D planes \cite{zhang2025fastlayout}.
In contrast, our Room Envelopes approach provides complete geometric details of the first layout surface for each pixel, rather than just 3D scene bounds.
This enables a more comprehensive scene reconstruction learning signal that supports scene completion from a single input image, treating layout estimation as a holistic scene completion task rather than only identifying the 3D extents of a room.

\section{The Room Envelopes Dataset}

In this section, we outline the properties of the Room Envelopes dataset, the construction procedure, and the data format and statistics.

\subsection{Hypersim Dataset}

The dataset is built upon the Hypersim dataset \cite{roberts2021hypersim}, which provides images of high-quality synthetic indoor environments, containing 77,400 images across 461 indoor scenes with detailed per-pixel annotations.
Hypersim includes images generated from fly-through trajectories of each scene, with multiple camera viewpoints per scene providing comprehensive coverage of each environment.
Each frame includes precise depth maps, surface normals, ground truth pointmaps providing world coordinates for each pixel location, instance-level semantic segmentations, and detailed lighting information that captures view-dependent effects such as glossy surfaces and specular highlights.
The scenes are rendered using V-Ray, a professional physically-based renderer that produces photorealistic imagery.
One limitation of this dataset is that the underlying 3D mesh assets are not freely available, and so they cannot be used as part of our pipeline.

\subsection{From Multi-view Images to Room Envelopes}

Our processing pipeline transforms the multi-view Hypersim data into Room Envelope representations through point cloud aggregation, semantic filtering, and view-specific re-rendering.
We extract pointmaps, surface normals, camera poses, and semantic segmentations from each frame in the Hypersim dataset, then aggregate these into point clouds.

Given pointmaps $\mathbf{P}_i \in \mathbb{R}^{H \times W \times 3}$, RGB images $\mathbf{C}_i \in \mathbb{R}^{H \times W \times 3}$, surface normals $\mathbf{N}_i \in \mathbb{R}^{H \times W \times 3}$, and semantic labels $\mathbf{S}_i \in \mathbb{Z}^{H \times W}$ for each frame $i$, we directly aggregate all world coordinate points with their associated attributes from all frames in the scene
\begin{equation}
\mathcal{P} = \bigcup_{i=1}^{N} \{(\mathbf{p}_{u,v}^{(i)}, \mathbf{c}_{u,v}^{(i)}, \mathbf{n}_{u,v}^{(i)}, s_{u,v}^{(i)}) : \mathbf{p}_{u,v}^{(i)} \in \mathbf{P}_i\},
\end{equation}
where $\mathbf{p}_{u,v}^{(i)} \in \mathbb{R}^3$ represents the world coordinate point, $\mathbf{c}_{u,v}^{(i)} \in \mathbb{R}^3$ the RGB colour, $\mathbf{n}_{u,v}^{(i)} \in \mathbb{R}^3$ the surface normal, and $s_{u,v}^{(i)} \in \mathbb{Z}$ the semantic label at pixel $(u,v)$ in frame $i$, with only valid points included.
We then downsample the pointcloud to remove redundant points using a voxel downsampling method with voxel size $\rho$, while preserving the associated attributes for each point.

Next, we filter the aggregated point cloud $\mathcal{P}$ to retain only layout-relevant points: walls, floors, doors, windows, and ceilings, such that
\begin{equation}
\mathcal{P}_\text{layout} = \{\mathbf{p} \in \mathcal{P} : s(\mathbf{p}) \in \mathcal{S}_\text{layout}\},
\end{equation}
where $s(\mathbf{p})$ denotes the semantic class of point $\mathbf{p}$ and $\mathcal{S}_\text{layout}$ is the set of layout-relevant semantic classes.
We align the filtered points to each original camera viewpoint by transforming them through a series of coordinate conversions to account for Hypersim's non-standard camera model
\begin{equation}
\tilde{\mathbf{p}}_\text{cam} = \mathbf{T}_\text{rend} \mathbf{T}_\text{w2c} \tilde{\mathbf{p}}_\text{w},
\end{equation}
where $\tilde{\mathbf{p}}$ represents points in homogeneous coordinates, $\mathbf{T}_\text{w2c}$ is the world-to-camera transformation matrix, and $\mathbf{T}_\text{rend}$ handles renderer-specific coordinate conversions and camera model corrections.
We then rasterize the scene by projecting the 3D points into pixel image space.
Since layout elements may project to the same pixel we isolate the closest layout surface using depth thresholding.
For each pixel $(u,v)$, we first identify all layout points that project to this pixel,
\begin{equation}
\mathcal{P}_{u,v} = \{\mathbf{p} \in \mathcal{P}_\text{layout} : \text{proj}(\mathbf{p}) = (u,v)\},
\end{equation}
where $\text{proj}(\mathbf{p})$ is the projection of 3D point $\mathbf{p}$ to image coordinates.
We filter these candidate points by depth proximity to select points near the closest surface 
\begin{equation}
\mathcal{P}_{u,v}^{\tau} = \{\mathbf{p} \in \mathcal{P}_{u,v} : |z_{\mathbf{p}} - z_{\text{min}}| \leq \tau\},
\end{equation}
where $z_{\mathbf{p}}$ is the camera-space depth (z-coordinate) of point $\mathbf{p}$, $z_{\text{min}} = \min_{\mathbf{p} \in \mathcal{P}_{u,v}} z_{\mathbf{p}}$, and $\tau$ is the depth threshold.
From this candidate set of points we select the point most aligned with the camera ray (the 3D ray from the camera centre through pixel (u,v)), minimising the distance between the point and its projection onto the ray.
This process is repeated for each pixel, resulting in a complete layout surface pointmap.

\begin{figure}[t]
    \centering
    \begin{tabular}{cc}
    \includegraphics[width=0.45\textwidth]{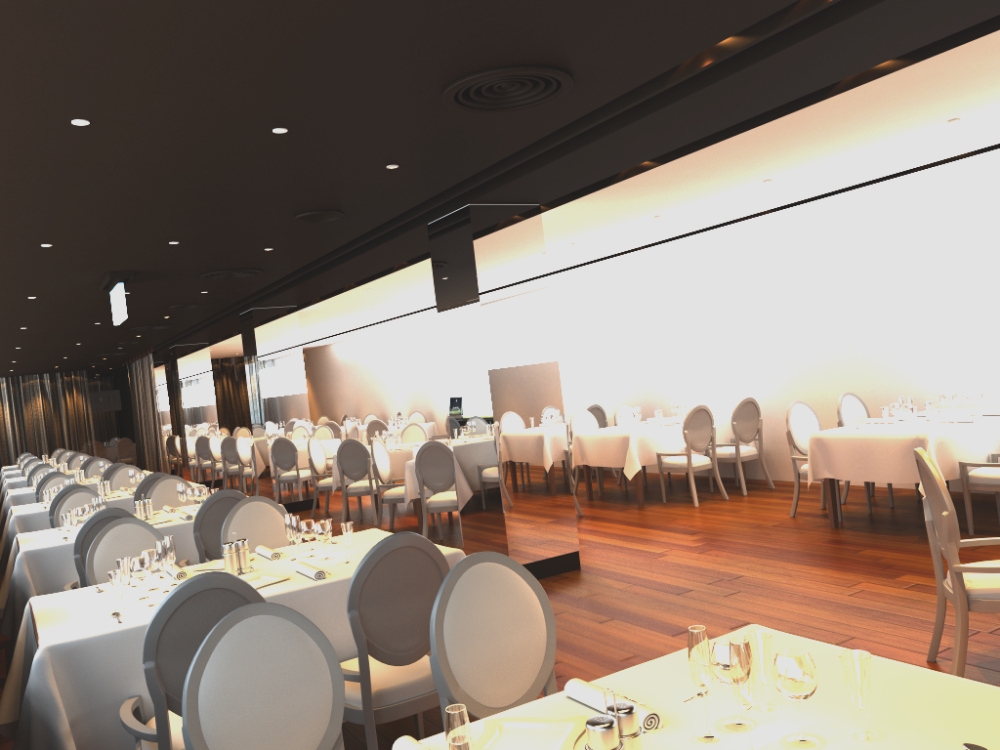} & 
    \includegraphics[width=0.45\textwidth]{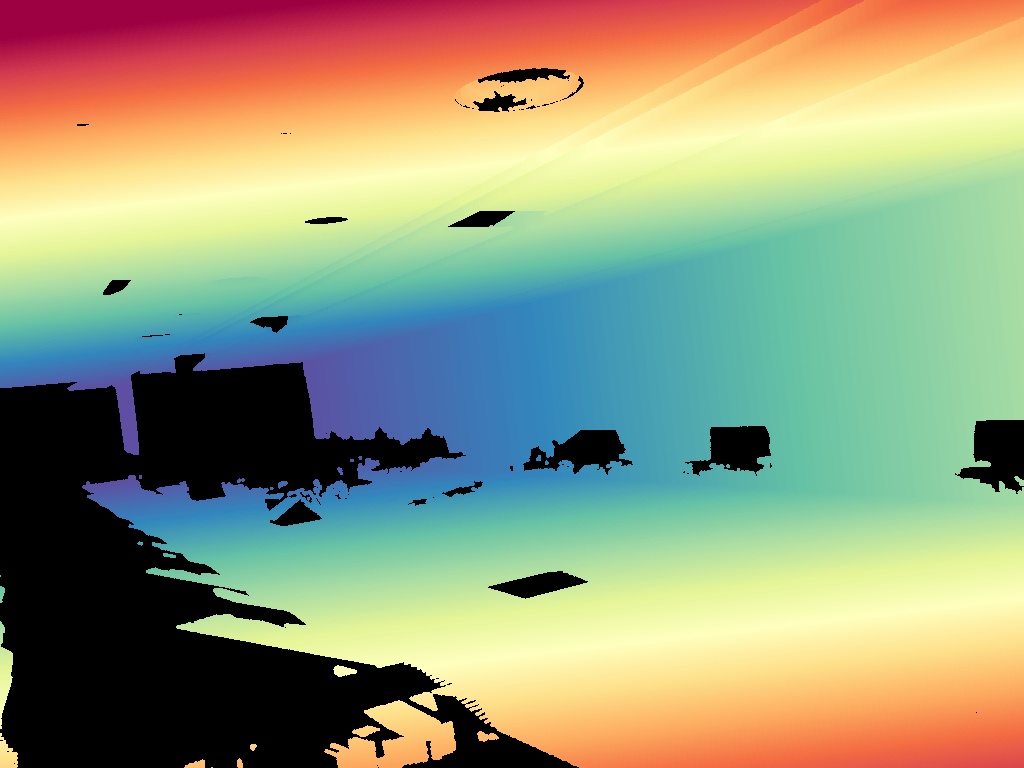} \\
    (a) RGB Image with occluded layout & (b) Surface layout with missing data \\
    \end{tabular}
    \caption{Missing data and occlusion patterns. (a) Original RGB image showing furniture occluding wall surfaces. (b) Corresponding layout depth image with holes (missing data) where no camera view captured the structural elements.}
    \label{fig:dataset_holes}
\end{figure}

A visualisation of the resulting room envelope is shown in \cref{fig:splash}.
It inherently contain gaps where layout elements are occluded across all viewpoints, for example, floor partially covered by a rug.
A visualisation highlighting these holes is shown in \cref{fig:dataset_holes}.
We observe that missing data of this type would also occur for real-world data collection, for which our processing pipeline could be used with minimal alterations.

\subsection{Dataset Format and Statistics}

The dataset comprises 77,400 images across 461 indoor scenes.
Each data sample consists of an RGB image paired with two pointmaps: a visible surface pointmap capturing all directly observable geometry, and a layout surface pointmap representing structural elements as they would appear without occlusion.
Additionally we retain metric scale point clouds of each scene, camera information, and surface normals per pixel.
Our approach maintains flexibility for future applications by preserving the completed point clouds, enabling extraction of alternative layers or conversion to other 3D representations as needed.

\section{Experiments}

\subsection{Experimental Setup}

We evaluate baseline methods on our Room Envelopes dataset using a comprehensive experimental setup designed to assess layout reconstruction performance. 
Our dataset is split into 80/10/10 train/validation/test partitions by scene, resulting in 365, 46, and 46 scenes respectively. 
This scene-level split ensures that test performance reflects the model's ability to generalise to unseen indoor environments rather than memorising specific room configurations.

We fine-tune a pre-trained MoGe v1 model as our base architecture, leveraging its established capabilities in single-view pointmap estimation. 
Training is conducted on 4 NVIDIA L40S 48GB GPUs with a batch size of 32, fine-tuning for 75,000 training steps over approximately 72 hours. 
We use the AdamW optimiser \cite{loshchilov2019decoupled} with a learning rate of $1 \times 10^{-6}$ for the backbone and $1 \times 10^{-5}$ for the prediction head, and train with the global pointcloud loss, mask loss, and normal loss from MoGe \cite{wang_moge_2024}.

\subsection{Quantitative Evaluation}

Our quantitative evaluation focuses on layout surface reconstruction accuracy across different visibility conditions. 
For comparison, we evaluate against two other methods. 
We compare against a pre-trained MoGe v1 model which estimates a pointmap from a single image \cite{wang_moge_2024}. 
Note that MoGe was trained on Hypersim among other datasets. 
We also compare against a pre-trained LaRI scene estimation model \cite{li2025lari}, which was trained on layered ray intersection images processed from the 3D-Front dataset \cite{3d_front}.
We concatenate the 5 layers of depth predictions from LaRI's output to create a pointcloud, and evaluate the quality of this pointcloud against our Room Envelopes dataset.

For fair comparison, predicted pointclouds are aligned to ground truth pointclouds using scale-shift alignment with least squares optimisation, where a single scale factor is applied to all coordinates and a 3D $z$ translation vector is added.
All pointclouds are scaled to metres using the scaling parameters provided in the Hypersim dataset \cite{roberts2021hypersim}. 
We evaluate reconstruction quality using Chamfer Distance (CD) and F-Scores (F) similar to previous works \cite{li2025lari}.
For LaRI evaluation, we take the best one-directional chamfer distance as the evaluation metric, since their method predicts many more points than in the ground truth.

Our evaluation distinguishes between ``seen'', ``unseen'', and ``overall'' layout regions. 
Seen regions correspond to layout elements that are fully visible in the source image, while unseen regions represent areas occluded by furniture and other objects but reconstructed using information from other viewpoints. 
Overall regions include all layout elements, regardless of visibility.
In our test set, unseen pixels comprise 23.57\% of all layout pixels, representing a substantial reconstruction challenge. 
The quantitative results are summarised in \cref{tab:layout_quantitative}, demonstrating our method's effectiveness particularly in reconstructing occluded layout geometry.

\begin{table*}[t]
    \centering
    \caption{Comparison of layout estimation performance on layout pixels seen in the input image, unseen in the input image, and seen or unseen in the input image (overall). CD denotes the chamfer distance and F denotes the F-score.}
    \footnotesize
    \setlength{\tabcolsep}{1pt}
    \begin{tabularx}{\linewidth}{@{}lCCCCCCCCC@{}}
    \toprule
    & \multicolumn{3}{c}{Seen} & \multicolumn{3}{c}{Unseen} & \multicolumn{3}{c}{Overall} \\
    \cmidrule(lr){2-4} \cmidrule(lr){5-7} \cmidrule(lr){8-10}
    Method & CD $\downarrow$ & F@0.1$\uparrow$ & F@0.05$\uparrow$ & CD$\downarrow$ & F@0.1$\uparrow$ & F@0.05$\uparrow$ & CD$\downarrow$ & F@0.1$\uparrow$ & F@0.05$\uparrow$ \\
    \midrule
    MoGe~\cite{wang_moge_2024} & \textbf{0.229} & \textbf{0.657} & 0.507 & 1.037 & 0.114 & 0.059 & \textbf{0.330} & 0.565 & 0.431 \\
    LaRI~\cite{li2025lari} & 0.948 & 0.412 & 0.286 & 1.522 & 0.400 & 0.283 & 0.925 & 0.410 & 0.285 \\
    Ours & 0.362 & 0.647 & \textbf{0.512} & \textbf{0.753} & \textbf{0.512} & \textbf{0.390} & 0.401 & \textbf{0.633} & \textbf{0.499} \\
    \bottomrule
    \end{tabularx}
    \label{tab:layout_quantitative}
\end{table*}

We evaluate the geometric consistency of our predicted layout surfaces by analysing how well normals in occluded regions conform to the distribution of normals from visible regions.
Normal vectors are computed from the estimated point cloud using local surface fitting.
We randomly sample 5,000 normal vectors from the seen layout regions to construct a kernel density estimator using von Mises--Fisher kernels, where each sampled normal serves as a kernel centre with equal weight, and concentration parameter $\kappa = 15$.
We then evaluate the likelihood of 5,000 randomly sampled normals in unseen layout regions under this kernel density estimator.
This process is repeated for each test image and the resulting likelihood values are averaged across the entire test split.
We compare against two baselines: uniformly distributed random normals on the unit sphere, and normals fixed to the most frequent direction in the seen regions of each test sample.
\cref{tab:normal_statistics} shows that our estimated normals in occluded regions are substantially more likely under the seen region distribution than uniformly random normals, though not as consistent as simply using the dominant normal direction from visible regions. 
This indicates that our method produces both geometrically plausible point distributions in occluded areas and maintains geometric variation rather than defaulting to a single dominant orientation.

\begin{table}[t]
    \centering
    \caption{Surface Normal Likelihood Analysis. We construct a kernel density estimator using von Mises--Fisher kernels from normals in seen layout regions and evaluate the likelihood of normals in unseen regions under this distribution. We compare our method against uniformly distributed random normals (Baseline-uniform) and normals fixed to the most frequent direction in seen regions (Baseline-dominant). Higher average likelihood values indicate more geometrically consistent normal predictions in unseen areas.
    }
    \begin{tabularx}{0.8\linewidth}{lC}
    \toprule
    Method & Average Likelihood $\uparrow$\\
    \midrule
    Baseline-uniform & 0.0795\\ 
    Baseline-dominant & 0.8788\\ 
    Ours & 0.4093\\
    \bottomrule
    \end{tabularx}
    \label{tab:normal_statistics}
\end{table}

\subsection{Qualitative Results}

\begin{figure*}[t]
    \centering
    \begin{tabular}{cccc}
    \textbf{Input RGB} & \textbf{Ground Truth} & \textbf{Ours} & \textbf{LaRI} \\
    \includegraphics[width=0.23\textwidth]{figures/ai_003_004-cam_01-11_image.jpg} & 
    \includegraphics[width=0.23\textwidth]{figures/ai_003_004-cam_01-11_depth.jpg} & 
    \includegraphics[width=0.23\textwidth]{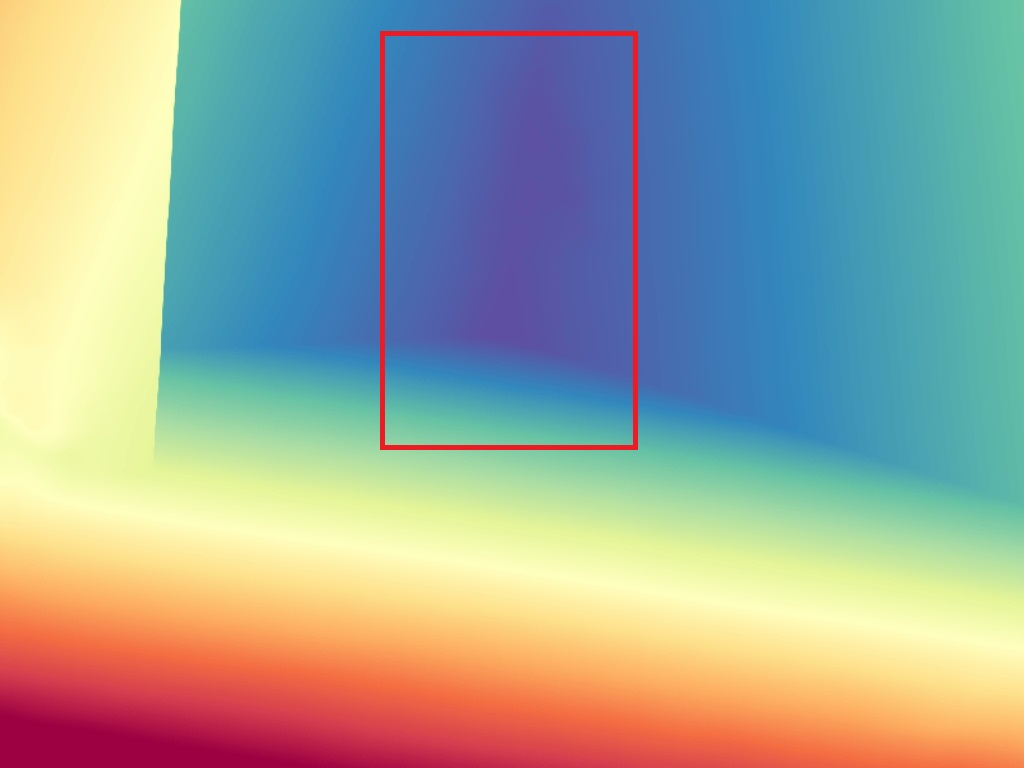} & 
    \includegraphics[width=0.23\textwidth]{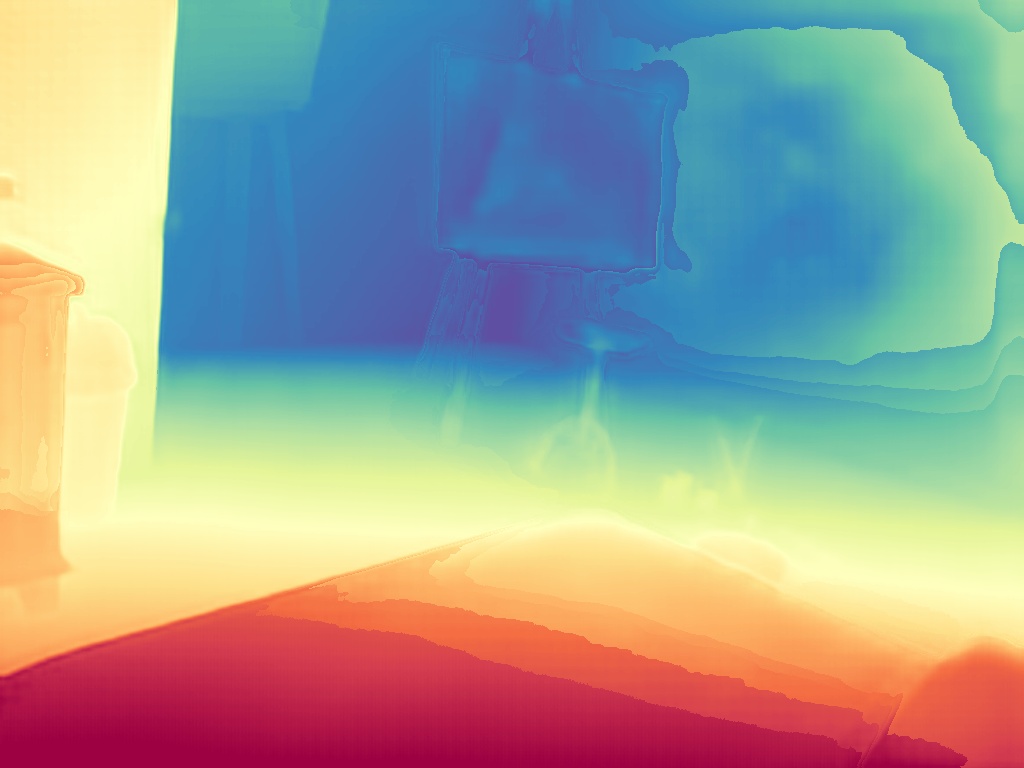} \\
    
    \includegraphics[width=0.23\textwidth]{figures/ai_017_003-cam_00-13-image.jpg} & 
    \includegraphics[width=0.23\textwidth]{figures/ai_017_003-cam_00-13_depth.jpg} & 
    \includegraphics[width=0.23\textwidth]{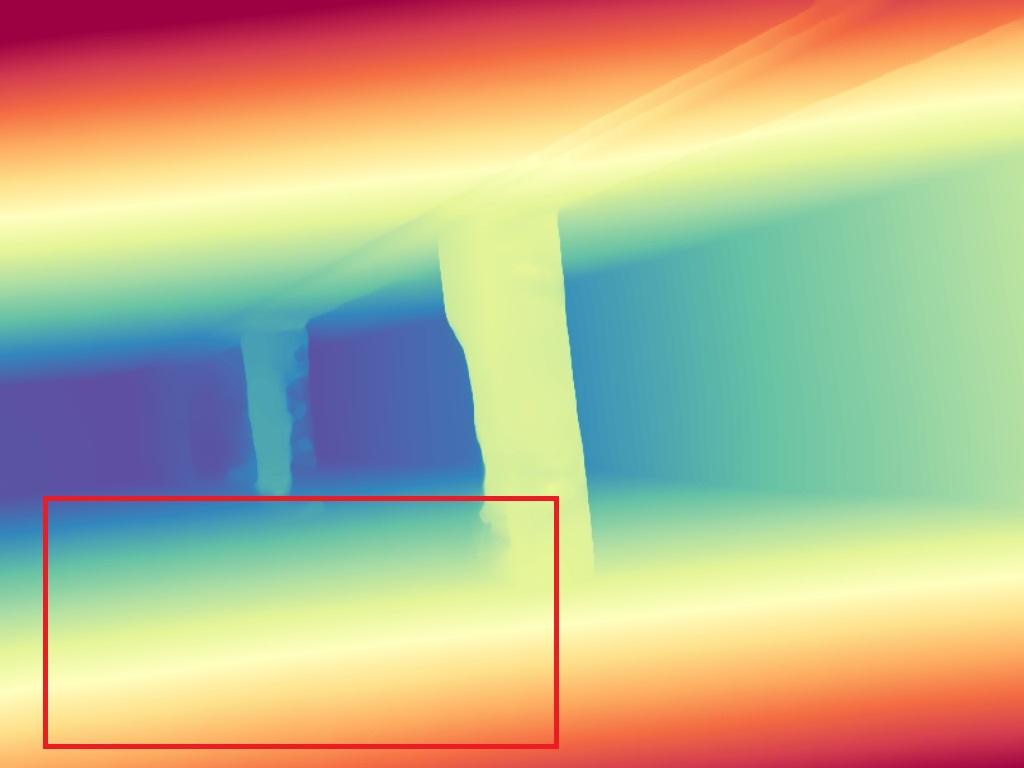} & 
    \includegraphics[width=0.23\textwidth]{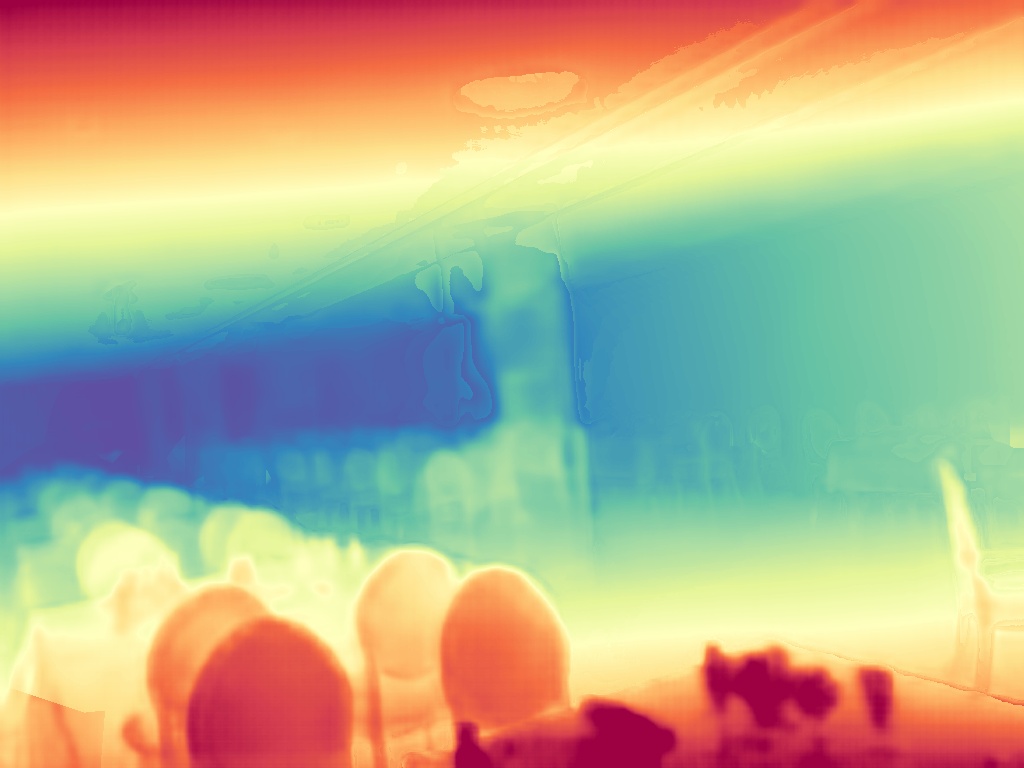} \\
    
    \includegraphics[width=0.23\textwidth]{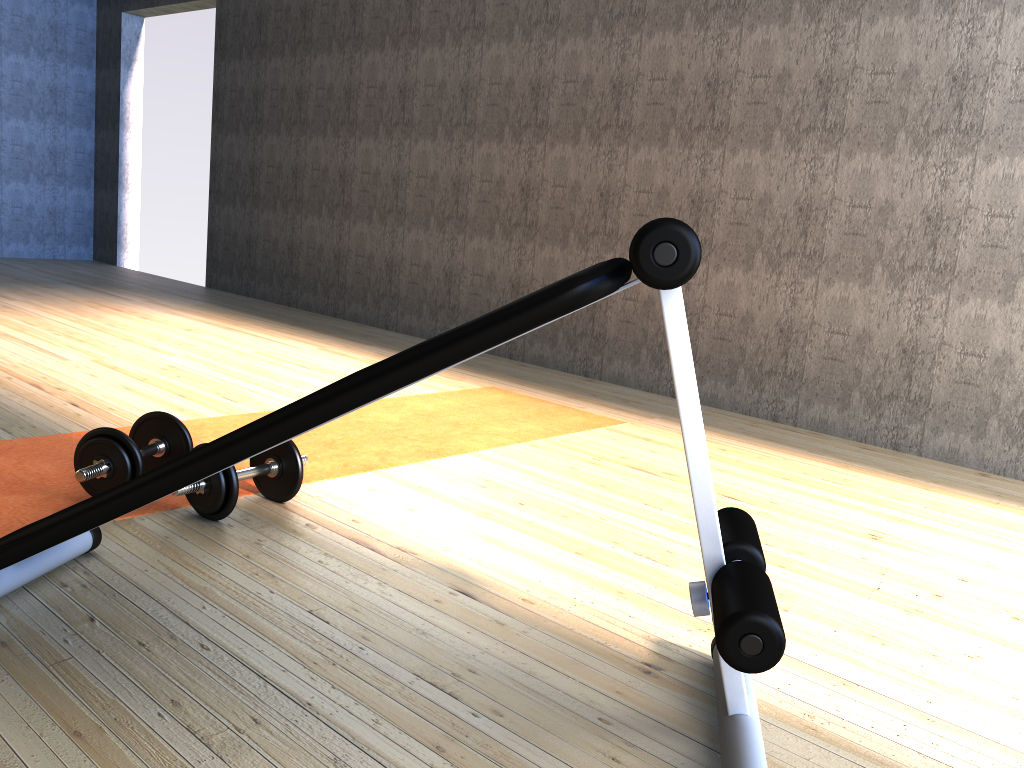} & 
    \includegraphics[width=0.23\textwidth]{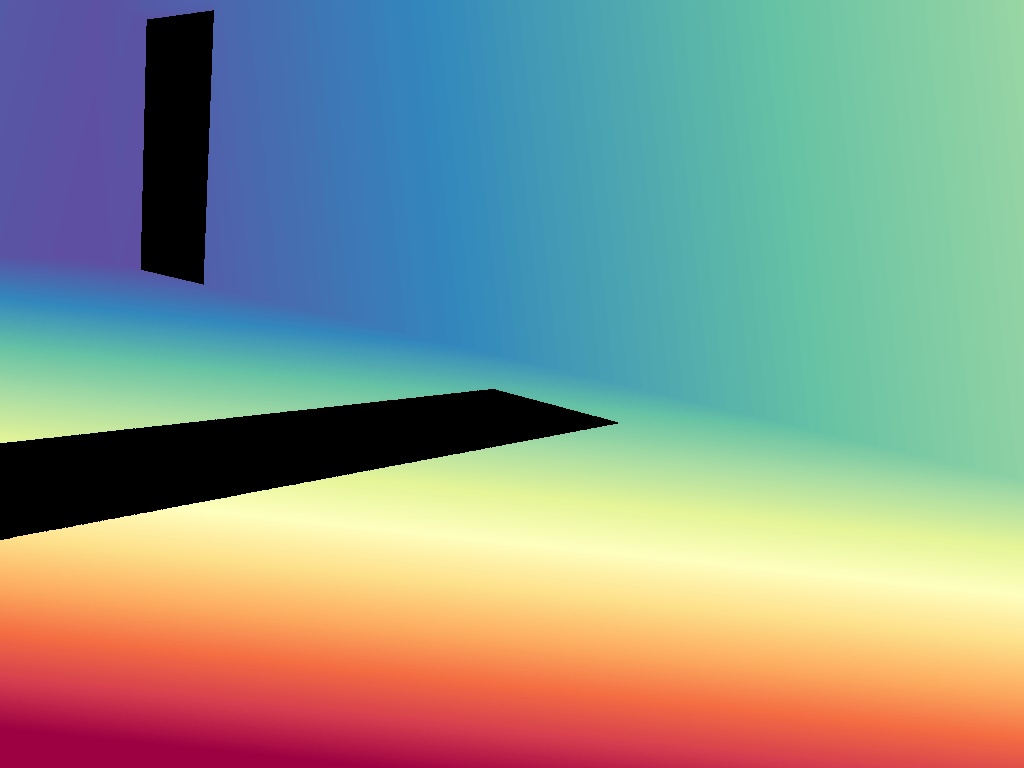} & 
    \includegraphics[width=0.23\textwidth]{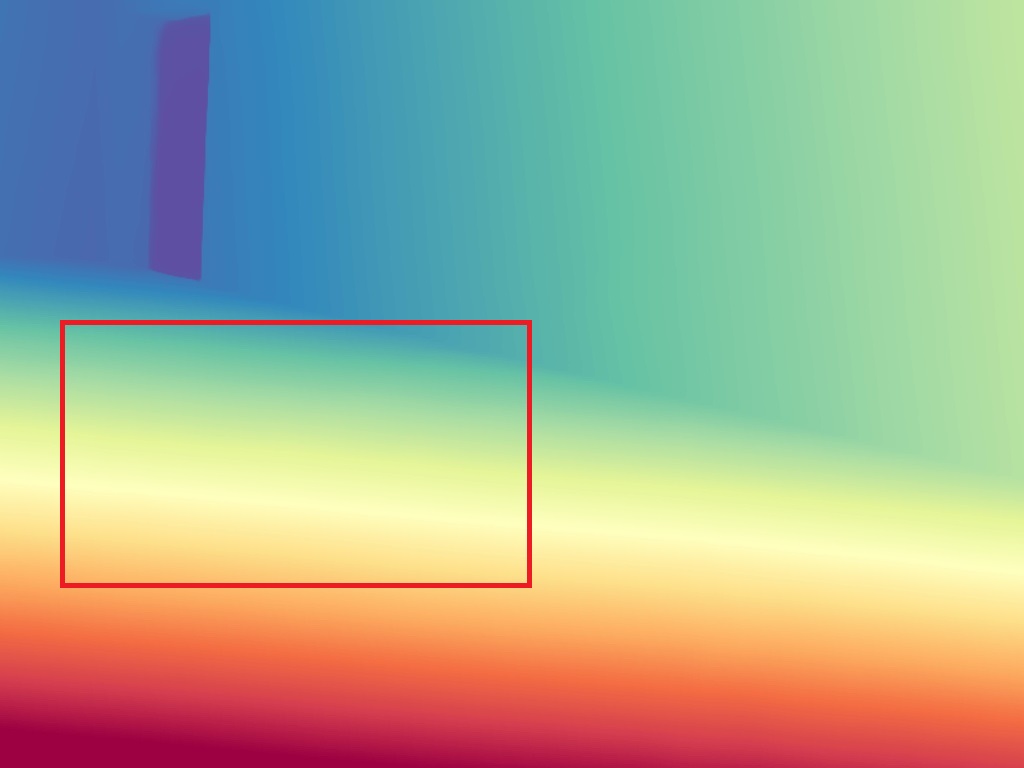} & 
    \includegraphics[width=0.23\textwidth]{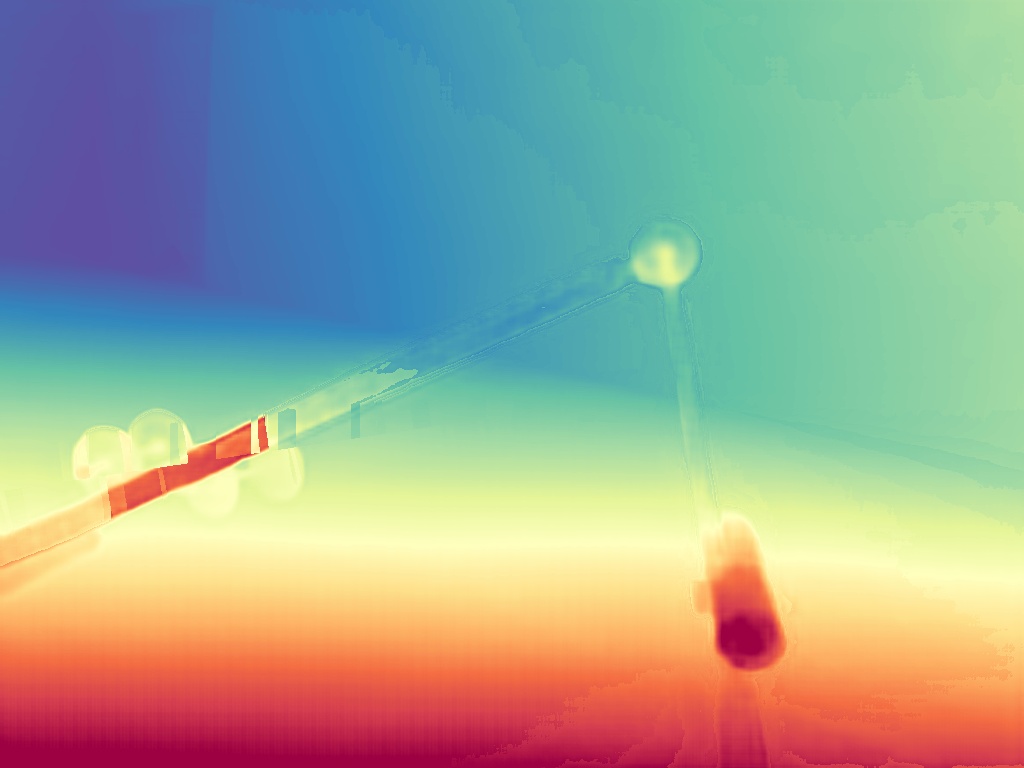} \\
    
    \includegraphics[width=0.23\textwidth]{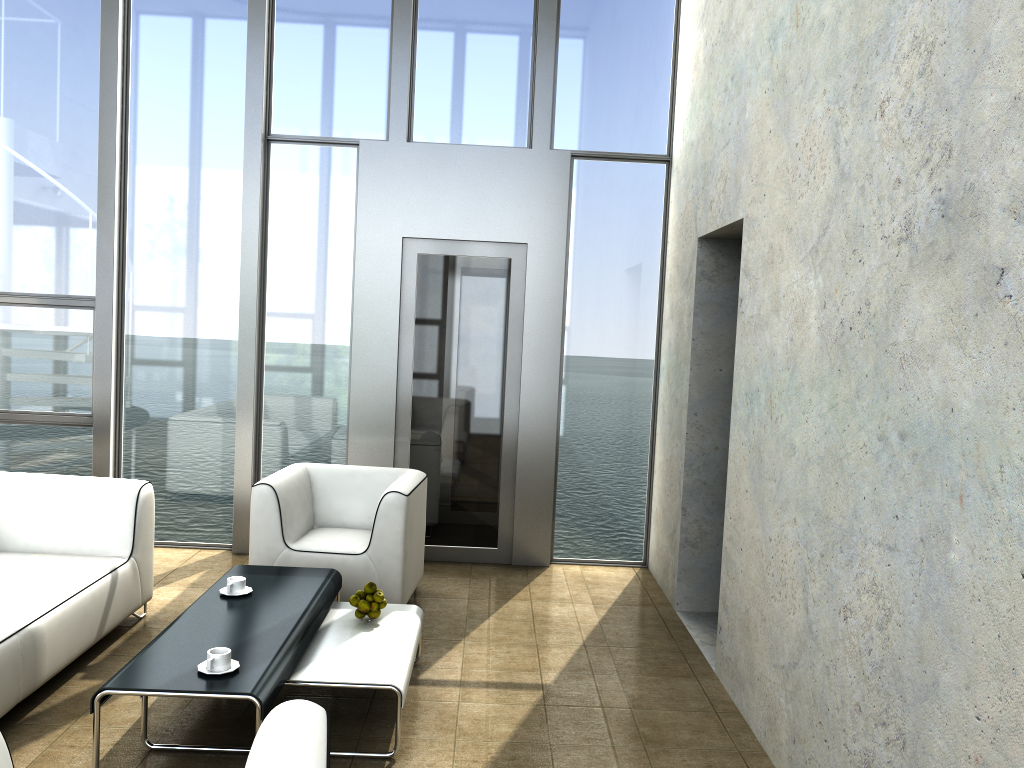} & 
    \includegraphics[width=0.23\textwidth]{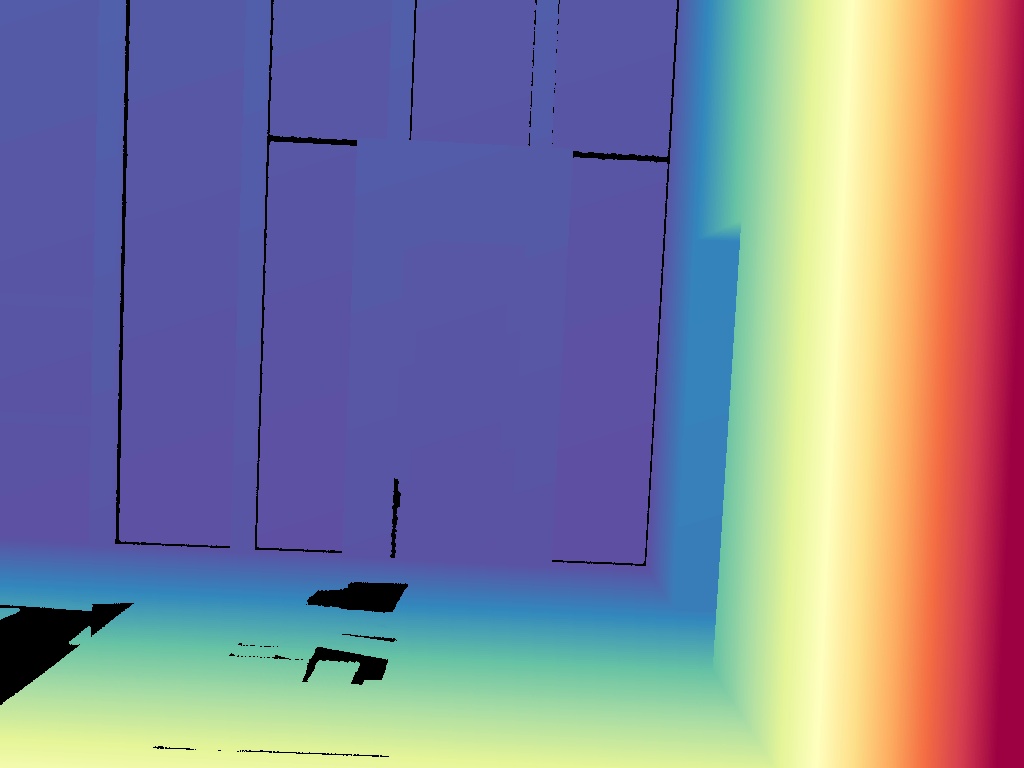} & 
    \includegraphics[width=0.23\textwidth]{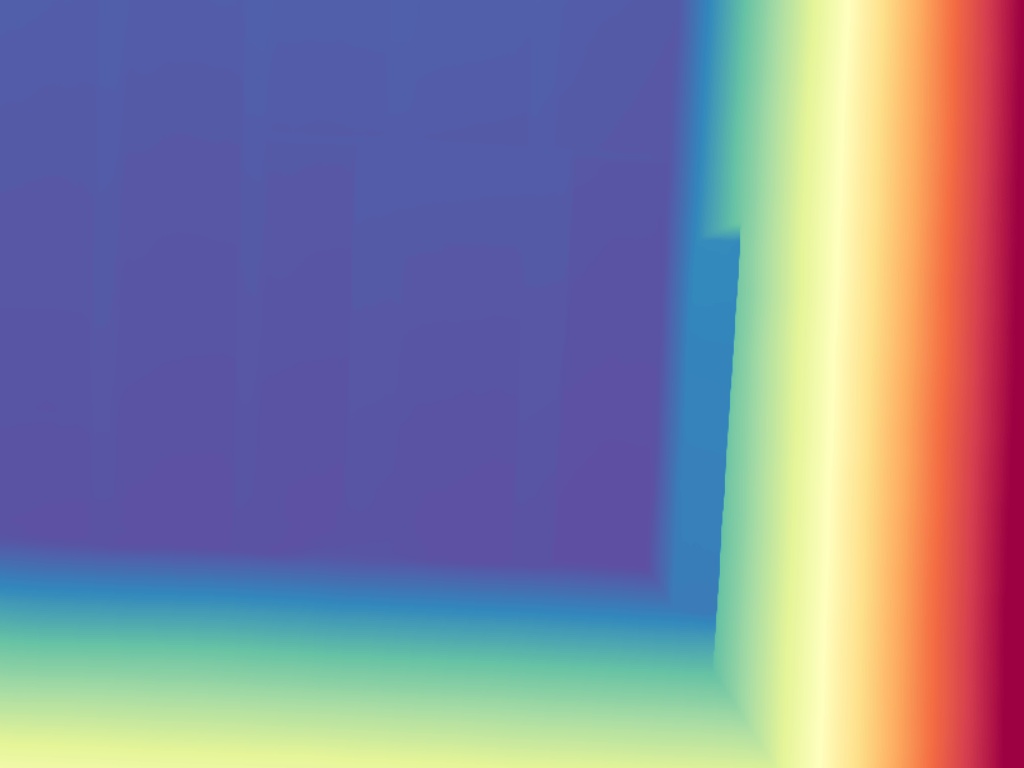} & 
    \includegraphics[width=0.23\textwidth]{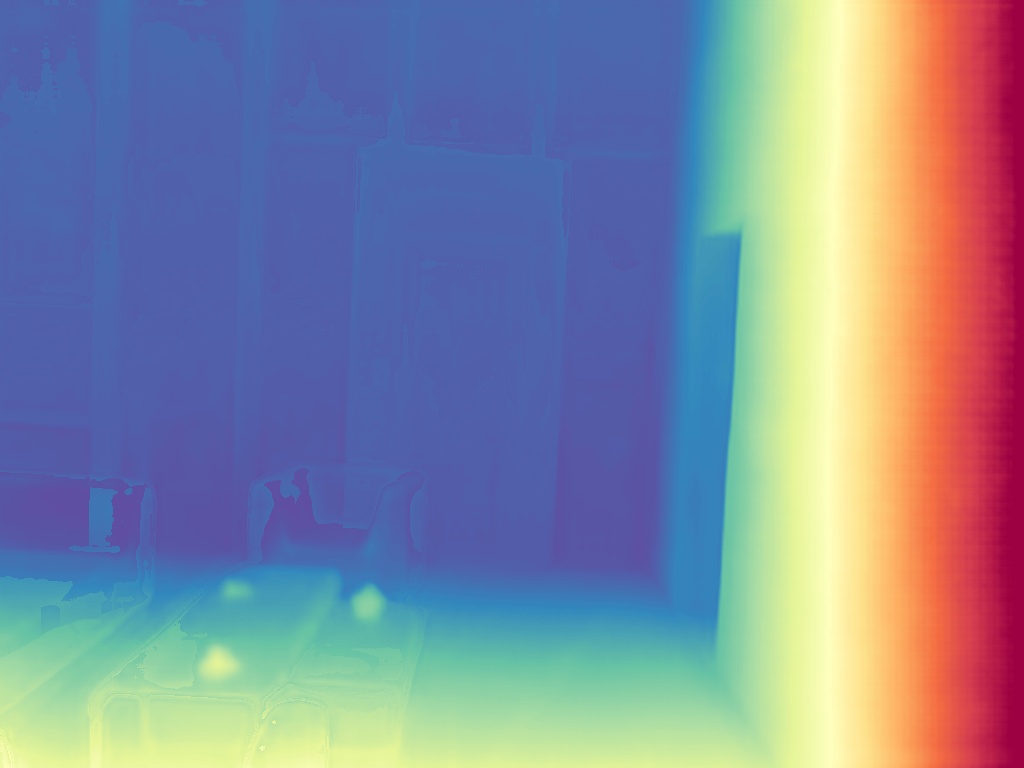} \\
    \end{tabular}
    \caption{Qualitative comparison of 3D first surface layout results as depth images. From left to right: input RGB image, ground truth layout geometry, our fine-tuned model trained on Room Envelopes, LaRI \cite{li2025lari}. Our method shows superior reconstruction of occluded layout elements. Red boxes highlight regions where our method successfully reconstructs layout geometry that is completely occluded in the input image.}
    \label{fig:qualitative_comparison}
\end{figure*}

\Cref{fig:qualitative_comparison} shows qualitative comparisons of our layout reconstruction against LaRI \cite{li2025lari} on several test set scenes. 
Our Room Envelopes approach demonstrates superior reconstruction of layout elements, particularly in occluded regions where furniture blocks the direct view of walls and structural elements. 
For LaRI, we visualise the best possible permutation of their 5 predicted depth images that best aligns to the ground truth. 
Additional in-the-wild qualitative results on real indoor images are provided in \cref{sec:wild_results}. 

\section{Conclusion}

We have introduced Room Envelopes, a synthetic dataset that addresses fundamental limitations in indoor scene reconstruction by providing dual pointmap representations: visible surfaces and layout surfaces. 
This dual representation enables direct supervision for layout reconstruction tasks, eliminating the ambiguity inherent in layered depth approaches and providing clear geometric targets for feed-forward models.
This work opens several directions for future research, including extension to complete scene geometry reconstruction within camera frustums, integration with real-world data collection pipelines, and application to downstream tasks such as robotic navigation and augmented reality.
\bibliographystyle{splncs04}
\bibliography{references}

\newpage
\appendix

\section{In-the-Wild Qualitative Results}
\label{sec:wild_results}
\begin{figure*}
    \centering
    \begin{tabular}{ccccc}
    \textbf{Source Image} & \textbf{MoGe Depth} & \textbf{Ours Depth} & \textbf{MoGe Normals} & \textbf{Ours Normals} \\
    \includegraphics[width=0.18\textwidth]{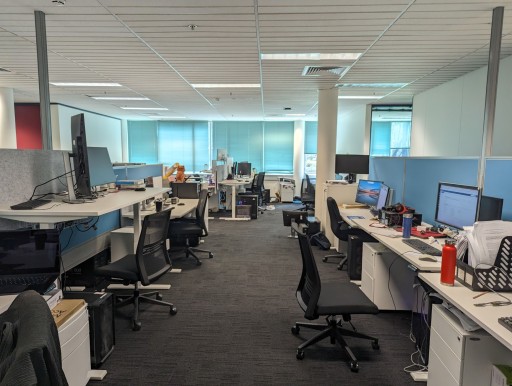} & 
    \includegraphics[width=0.18\textwidth]{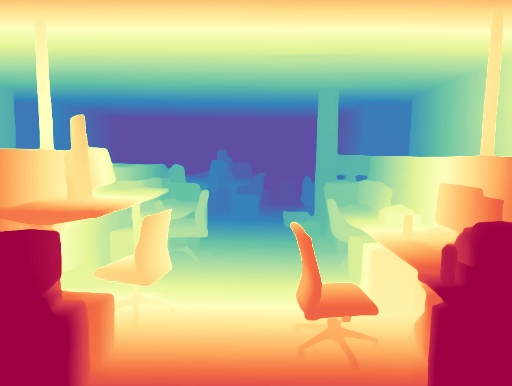} & 
    \includegraphics[width=0.18\textwidth]{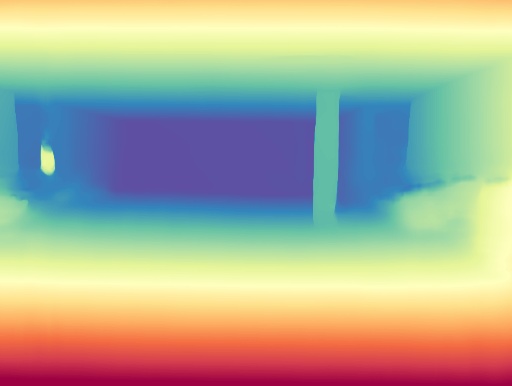} & 
    \includegraphics[width=0.18\textwidth]{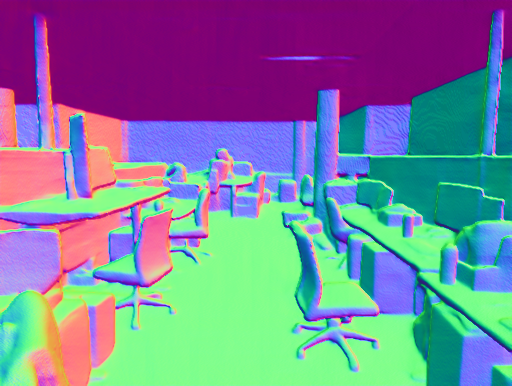} & 
    \includegraphics[width=0.18\textwidth]{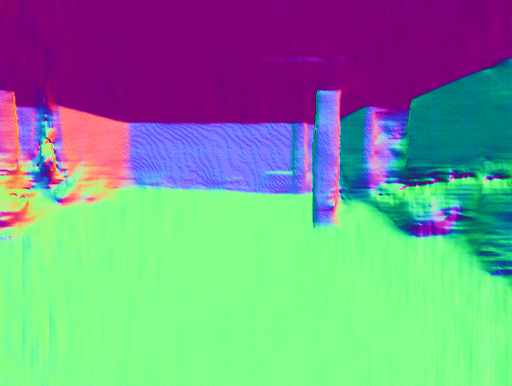} \\
    
    \includegraphics[width=0.18\textwidth]{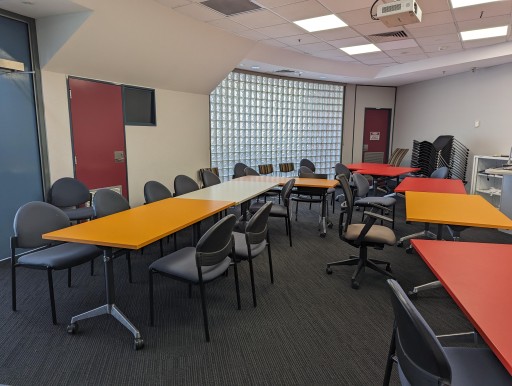} & 
    \includegraphics[width=0.18\textwidth]{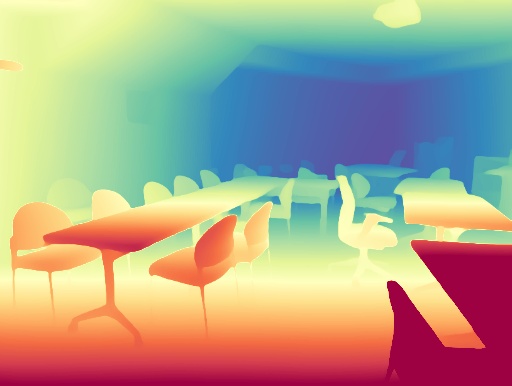} & 
    \includegraphics[width=0.18\textwidth]{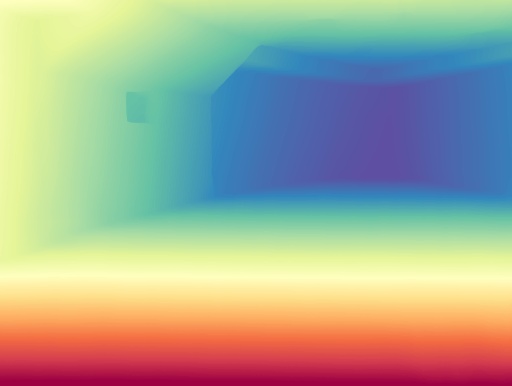} & 
    \includegraphics[width=0.18\textwidth]{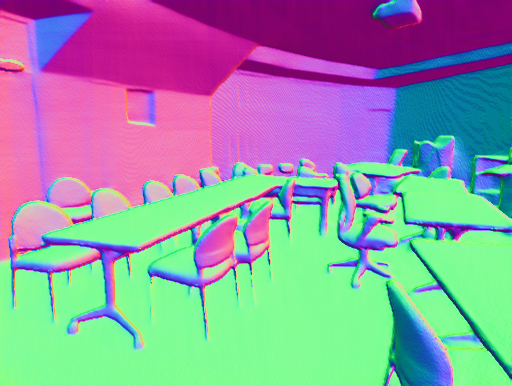} & 
    \includegraphics[width=0.18\textwidth]{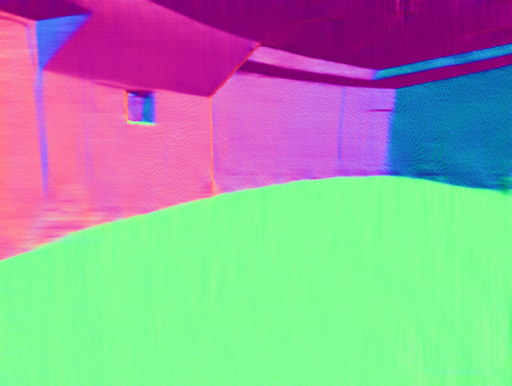} \\
    
    \includegraphics[width=0.18\textwidth]{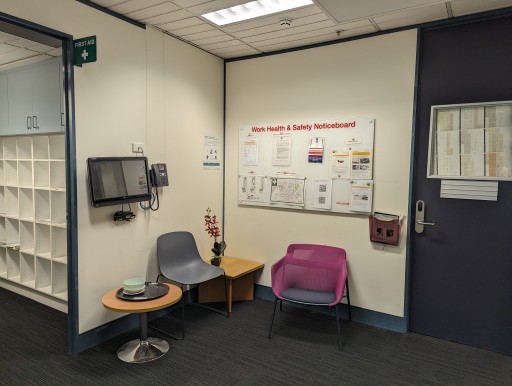} & 
    \includegraphics[width=0.18\textwidth]{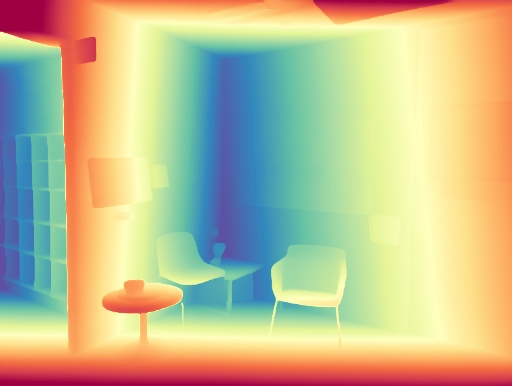} & 
    \includegraphics[width=0.18\textwidth]{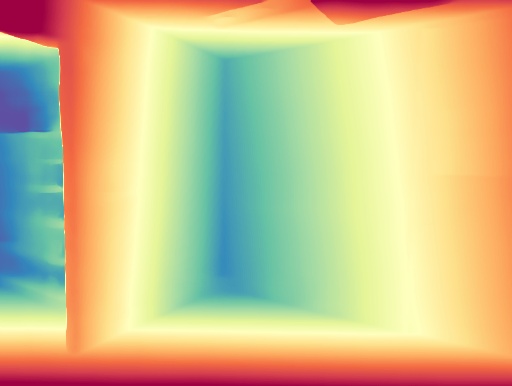} & 
    \includegraphics[width=0.18\textwidth]{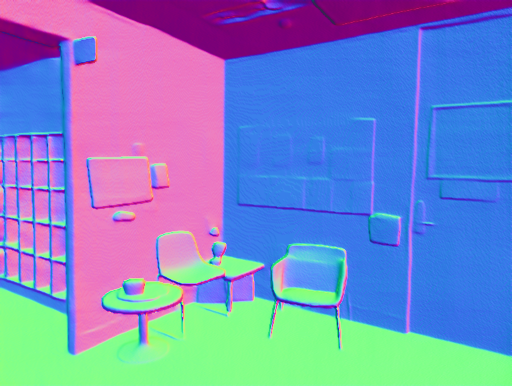} & 
    \includegraphics[width=0.18\textwidth]{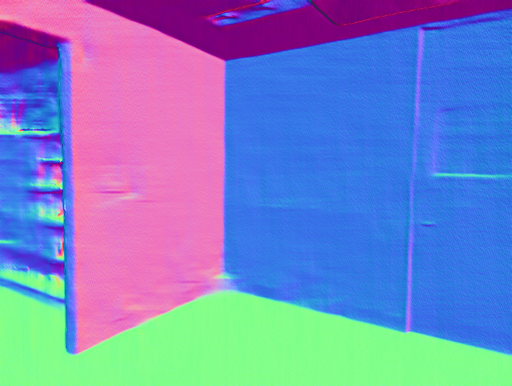} \\
    
    \includegraphics[width=0.18\textwidth]{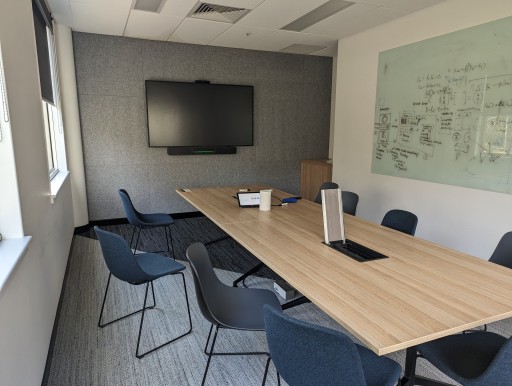} & 
    \includegraphics[width=0.18\textwidth]{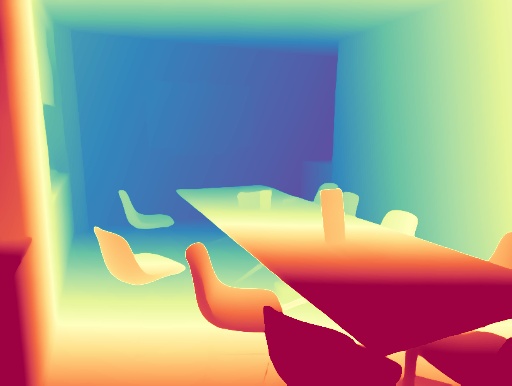} & 
    \includegraphics[width=0.18\textwidth]{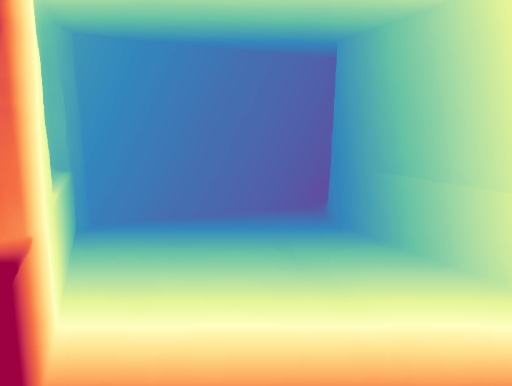} & 
    \includegraphics[width=0.18\textwidth]{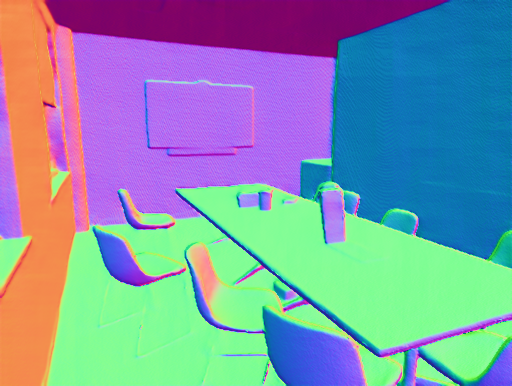} & 
    \includegraphics[width=0.18\textwidth]{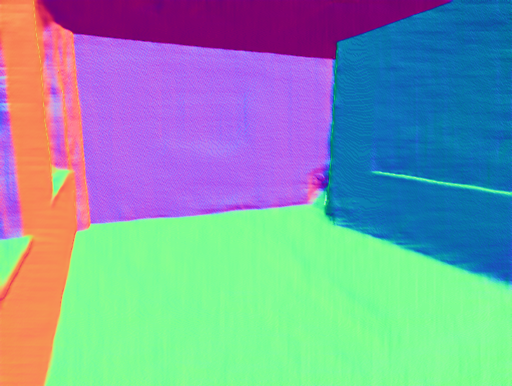} \\
    \end{tabular}
    \caption{Qualitative comparison on in-the-wild images captured by a phone camera comparing MoGe and our layout trained model on real indoor images. Normals are estimated using local surface fitting and colourised by mapping the x, y, z components to red, green, blue channels respectively. Please zoom in to see finer details.}
    \label{fig:wild_results}
\end{figure*}

We present additional qualitative results on real-world indoor images captured from indoor environments in \cref{fig:wild_results}.
These examples show how our layout estimation model performs on real indoor environments with different lighting conditions and realistic furniture arrangements.
We convert the estimated pointmap into a depth image, and estimate surface normals through local surface fitting on the pointmap.
The results illustrate the model's ability to reconstruct plausible layout geometry even when applied to real images that differ from the synthetic training data, highlighting the value of our layout supervision approach for practical applications.

\end{document}

%% file: macros.tex
\newcommand{\xmark}{\ding{55}}